\documentclass{article}
\usepackage{spconf,amsmath,graphicx}
\usepackage{booktabs}
\usepackage{array}
\usepackage{balance}
\usepackage{bibspacing}
\balance
\usepackage[pagebackref,breaklinks,colorlinks]{hyperref}
\usepackage[capitalize]{cleveref}
\crefname{section}{Sec.}{Secs.}
\Crefname{section}{Section}{Sections}
\Crefname{table}{Table}{Tables}
\crefname{table}{Tab.}{Tabs.}

\title{Patch-wise Auto-Encoder for Visual Anomaly Detection}
\name{Yajie Cui, Zhaoxiang Liu*\thanks{*Corresponding author}, Shiguo Lian*}
\address{Unicom Digital Technology, China Unicom, Beijing 100013, China}

\begin{document}
%
\maketitle
\begin{abstract}
Anomaly detection without priors of the anomalies is challenging. In the field of unsupervised anomaly detection, traditional auto-encoder (AE) tends to fail based on the assumption that by training only on normal images, the model will not be able to reconstruct abnormal images correctly. On the contrary, we propose a novel patch-wise auto-encoder (Patch AE) framework, which aims at enhancing the reconstruction ability of AE to anomalies instead of weakening it. Each patch of image is reconstructed by corresponding spatially distributed feature vector of the learned feature representation, i.e., patch-wise reconstruction, which ensures anomaly-sensitivity of AE. 
Our method is simple and efficient. It advances the state-of-the-art performances on Mvtec AD benchmark, which proves the effectiveness of our model. It shows great potential in practical industrial application scenarios.
\end{abstract}
\vspace{-0.2cm}
\section{Introduction}
\vspace{-0.2cm}
\label{sec:introduction}
Aim of unsupervised anomaly detection is to identify abnormal images only based on knowledge of normal images. It is crucial especially in the intelligent manufacturing process that screens defective items and performs automatic inspections to guarantee that products are qualified. In the actual production line, defect rate is often relatively low. It may not be possible to collect sufficient quantities of defective samples. In addition, because of the various anomalies that can occur during manufacturing, a detection model trained on only a limited number of abnormal samples may not generalize to those that have not been seen before. So in practice, it is frequently preferable in these inspection circumstances to train models to detect abnormality using only normal images.

There exists some research in the field of unsupervised anomaly detection. Some methods \cite{RudolphWR19}\cite{autoencoder2014}\cite{BergmannLFSS19} conduct encoding and decoding on normal images and train the neural networks with the aim of reconstruction. The differences between the images before and after reconstruction are analyzed to detect anomalies in the detection stage. Some methods \cite{rudolph2021same}\cite{gudovskiy2022cflow}\cite{rezende2015variational} use normalizing flow to learn transformations between data distributions and well-defined densities. Their special property is that their mapping is bijective and they can be evaluated in both directions. It can be served as a suitable estimator of probability densities for the purpose of detecting anomalies. Some methods \cite{DefardSLA20}\cite{roth2021towards} extract meaningful vectors describing the entire image, and the anomaly score is usually represented by the distance between the embedded vectors of the test images and the reference vector representing normality from the training dataset. 

Among the methods based on reconstruction, one strategy to address anomaly detection problem is to exploit convolutional auto-encoders \cite{bergmann2019mvtec} \cite{MeiYY18}. Typical auto-encoder (AE) methods train the model based on normal samples to have the ability to reconstruct normal samples. In the test phase, abnormal samples are fed into the well trained AE, which should also be reconstructed into normal samples. So anomaly detection can be realized by comparing pixel-wise differences between the input and its reconstructed version via some distance metric, e.g. L2-distance or structural similarity metric (SSIM) \cite{BergmannLFSS19}. 
Typical AE methods function based on the assumption that by training only on normal images, the model will not be able to reconstruct abnormal images correctly, leading to higher anomaly scores.
\begin{figure*}[t]
\setlength{\abovecaptionskip}{-2cm} 
\setlength{\belowcaptionskip}{-2cm} 
  \centering
  \includegraphics[width=0.6\linewidth]{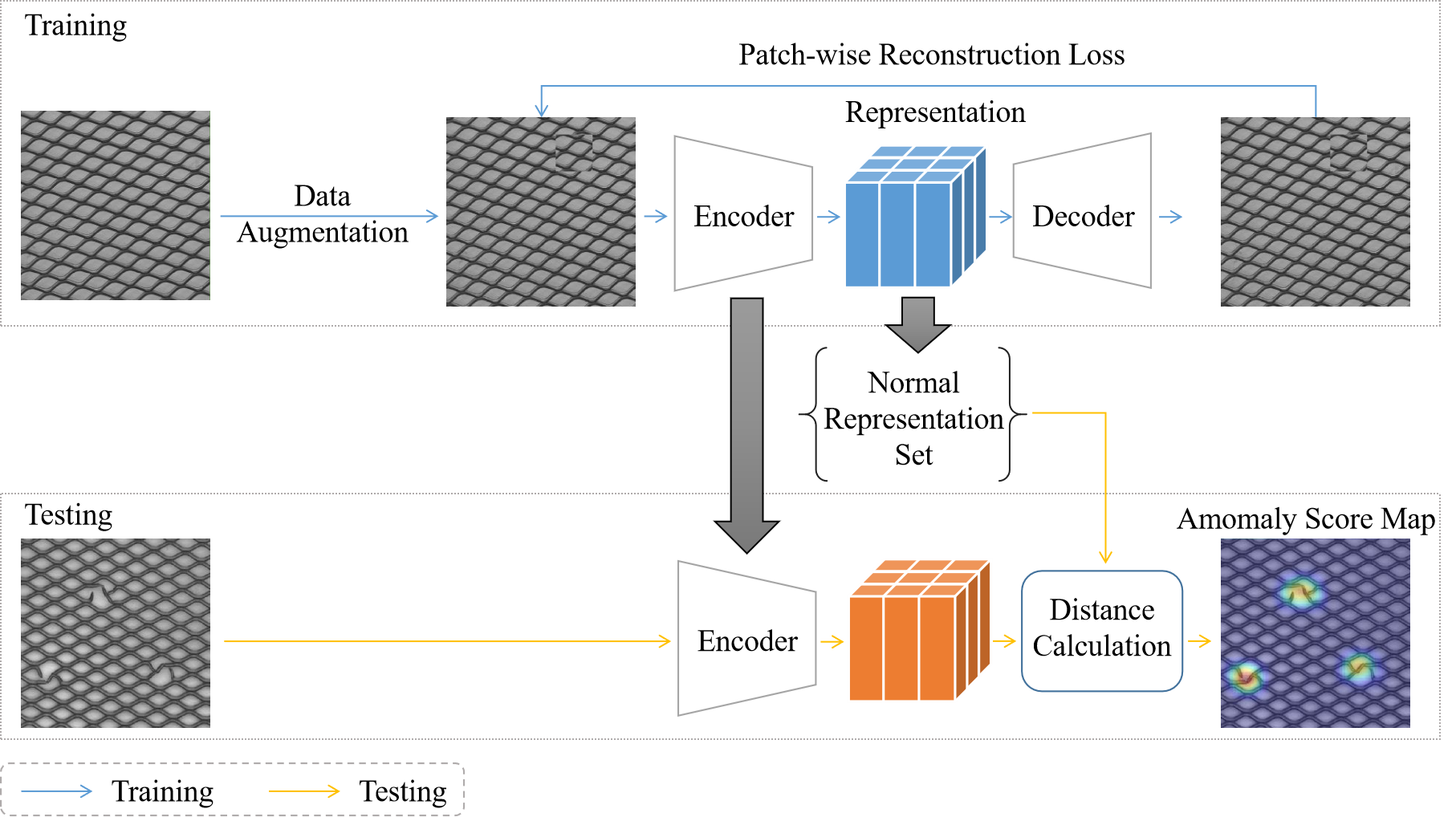}
   \caption{The overview of our Patch-wise Auto-Encoder anomaly detection pipeline.}
   \label{fig:framework}
\end{figure*}
The insensitiveness of AE to anomalies is the most important premise of AE based algorithms. While due to the strong generalization of AE, defects might also be well reconstructed. As a result, anomaly cannot be detected when the reconstruction is compared with the original image. Methods RIAD \cite{ZavrtanikKS21} and ITAD \cite{PirnayC22} struggle to solve this problem. The failure of AE for anomaly detection implies, to some degree, the latent representation of AE might be able to capture the anomaly patterns of input image. There are also other works \cite{gong2019memorizing} \cite{park2020learning} besides industrial field showing that AE has the ability to express the characteristics of abnormal samples. Inspired by this, we propose a method on the contrary to typical AE based methods, as shown in \cref{fig:framework}. Our idea attempts to enhance the reconstruction ability of AE to anomalies instead of weakening it during training. Enlightened by Cutpaste \cite{LiSYP21}, we augment the input normal samples with some artificial defects randomly. Consequently, the model can learn the pattern of normal samples during training stage, in the mean time, it can also learn the pattern of abnormal samples through the training of artificial defect samples, so as to improve the model's ability to reconstruct the original image and enhance its sensitivity to the details of the original image, especially the defective parts. In this way, our AE model could reconstruct not only normal images but also anomalous images very well. 

In addition to enhancing the model's encoding ability of defect information by enriching the input content, we also further promote the reconstruction ability of the intermediate overall representation. In the training process of the model, the intermediate representation reconstruct the original input by decoder. Different from typical image-wise reconstruction, patch-wise reconstruction is adopted in our framework. The overall feature representation, which is a three dimensional tensor, obtained by encoder retains spatial information of the original image. In the reconstruction process, each patch of the original image is reconstructed by corresponding spatially distributed feature vector of the tensor, so that each spatial feature vector has the ability to reconstruct each fine-grained image patch independently, without disturbance from other feature vectors. We notice that Patch SVDD \cite{YiY20} also conducts anomaly detection based on the idea of patch. But it explicitly divides the image into multiple patches and inputs them into the model in turn. Repeated feature extraction for each patch consumes more time. Instead, we input the whole image once into model and get spatially distributed feature vector. Each spatial feature vector corresponds to a region of image, which is called a patch. Consequently, our method can extract all the features at once.

We aim to obtain an overall feature representation sensitive to defects while keeping good representation ability to normality. This representation is further equipped with fine-grained image reconstruction capability through patch-wise decoding procedure. Then downstream anomaly detection task is carried out in the feature space instead of pixel space as typical AE methods do. It advances the state-of-the-art performances on Mvtec AD benchmark, which demonstrates the effectiveness of our approach.
\vspace{-0.2cm}
\section{Method}
\label{sec:method}
\begin{figure*}[t]
\setlength{\abovecaptionskip}{-2cm} 
\setlength{\belowcaptionskip}{-2cm} 
  \centering
  \includegraphics[width=0.7\linewidth]{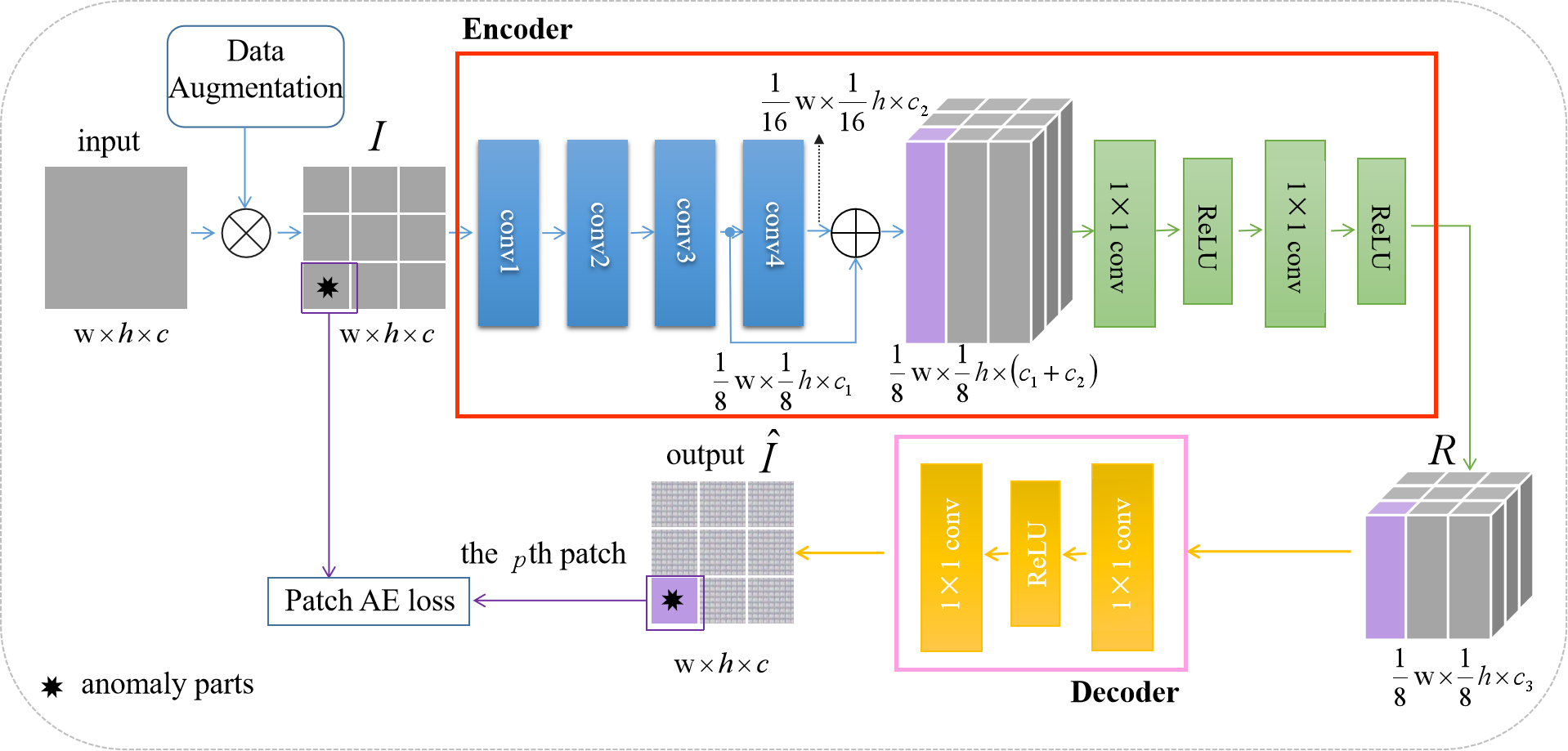}
   \caption{The detail of Patch AE structure, consisting of encoder and decoder, which are marked in red and pink boxes, respectively. $\bigotimes$ means performing data augmentation operation on the input image. $\bigoplus$ means feature concatenating after unifying dimension. Blue rectangles represents network layers initialized with pre-trained parameters. Green ones are two $1 \times 1$ convolution - ReLU activation layers of encoder. Yellow ones are $1 \times 1$ convolution, activation and $1 \times 1$ convolution layers of decoder. Purple annotations indicate the correspondence between feature vector and reconstructed image patch.}
   \label{fig:encoder}
\end{figure*}
\vspace{-0.2cm}
\subsection{Data Augmentation}
\vspace{-0.2cm}
Traditional AE method reconstructs abnormal samples into normal samples, resulting in difference between them, which is used for anomaly detection. However, this approach tends to fail because the differences in AE reconstruction are not very significant, which indicates that AE has the ability to express the characteristics of abnormal samples. Therefore, we come up with strengthening AE's ability to capture characteristics of abnormal sample, so as to equip the learned representation with stronger representation ability, which can not only describe normality but also cover abnormality information. In this way, the representation ability of feature representation and the information covered are better than that of the original image, leading to higher accuracy of downstream anomaly detection task in feature space.

In order to make the representation we learned sensitive to defects, we need to input some anomaly samples into the model to train its ability of capturing anomaly information. We borrowed the idea of Cutpaste \cite{LiSYP21} that synthesize some defects with different shapes, positions, angles and sizes to the defect-free samples. Whether to add defects, size, shape, angle and position of defects are all hyperparameters to tune for data augmentation.
\vspace{-0.3cm}
\subsection{Architecture of Patch AE}
\vspace{-0.2cm}
As shown in \cref{fig:framework}, the model is roughly an auto-encoder structure. In the training phase, we first input normal samples for data augmentation. Data after augmentation procedure consists of original normal samples and artificial abnormal samples. They are all input into encoder to obtain the intermediate feature representation, which is a three dimensional tensor retaining spatial information. Then each feature vector of the representation is fed into the decoder to reconstruct corresponding image patch.
In this way, the encoder has good ability to restore original image, guaranteeing the intermediate features obtained sensitive to anomaly. Consequently, the well trained encoder and the normal representation set extracted from normal samples can be obtained after training. 
During the testing phase, the test sample is input into the trained encoder to obtain the representation, which is used to calculate distance with normal representation set in feature space to get the anomaly score map.
\vspace{-0.2cm}
\subsection{Encoding Module}
\vspace{-0.2cm}
As shown in red box of \cref{fig:encoder}, our encoder consists of 4 convolutional layers, followed by a feature concatenating layer, two $1 \times 1$ convolution - ReLU activation layers. 
As is proven in other methods \cite{LiSYP21} \cite{YiY20} that training networks from scratch doesn't work well, in order to achieve better performance, higher training accuracy and faster convergence, we use pre-trained Wideresnet101 to initialize our first 4 convolutional layers. Extracting multi-scale features enables the learned representation to better cover both high-level semantic features and low-level features. So our model utilize several feature layers of different scales for fusion. The feature output by conv4 layer is upsampled to the same size as that of conv3, and the two features are then concatenated. The concatenated representation contains redundant information. Specifically, in addition to information of corresponding image patch, information of neighboring patches is also contained in each spatially distributed vector. The redundant information might harm distance metric of feature vector for anomaly detection. In order to eliminate redundancy and extract actual effective information from corresponding image patch, two $1 \times 1$ convolution - ReLU activation layers are used for dimension compression ($c_1+c_2>c_3$) of spatially distributed feature vectors to obtain an overall multi-scale fusion feature \textbf{R} of the image, which corresponds to the whole image \textbf{I}. The representation \textbf{R} covers effective information of features of different scales, and can fully express the semantic information and graphic information of the image.
\begin{table*}[]
\centering
\caption{Results for the task of anomaly detection on the MVTec dataset. Results are listed as AUROC scores and are marked individually for each class. An average score over all classes is reported in the last row.}
\label{table:Experiment result table.}
\resizebox{\textwidth}{!}{%
\begin{tabular}{@{}lcccccccccccc@{}}
\toprule
           & L2-AE \cite{BergmannLFSS19}& SSIM-AE \cite{BergmannLFSS19}& VAE-grad \cite{DehaeneFCE20}& RIAD \cite{ZavrtanikKS21}& CutPaste \cite{LiSYP21}& CS-Flow \cite{RudolphWRW22}& SSPCAB \cite{RisteaMINKMS22}& AST \cite{RudolphWRW23}& CFA \cite{LeeLS22}& Patchcore-single \cite{RothPZSBG22}& Patchcore-multi \cite{RothPZSBG22}& Patch AE \\ \midrule
Bottle     & 86    & 93      & 92       & 99.9  & 98.3     & 99.8    & 98.4   & 100   & 100   & 100      &- & 100            \\
Cable      & 86    & 82      & 91       & 81.9  & 80.6     & 99.1    & 96.9   & 98.5  & 99.8  & 99.6    &-  & 99.56          \\
Capsule    & 88    & 94      & 92       & 88.4  & 96.2     & 97.1    & 99.3   & 99.7  & 97.3  & 98.2     &- & 99.1           \\
Carpet     & 59    & 87      & 74       & 84.2  & 93.1     & 100     & 98.2   & 97.5  & 97.3  & 98.4     &- & 99.87          \\
Grid       & 90    & 94      & 96       & 99.6  & 99.9     & 99      & 100    & 99.1  & 99.2  & 99.8    &-  & 99             \\
Hazelnut   & 95    & 97      & 98       & 83.3  & 97.3     & 99.6    & 100    & 100   & 100   & 100      &- & 100            \\
Leather    & 75    & 78      & 93       & 100   & 100      & 100     & 100    & 100   & 100   & 100     &-  & 100            \\
Metal Nut  & 86    & 89      & 91       & 88.5  & 99.3     & 99.1    & 100    & 98.5  & 100   & 100      &- & 99.8           \\
Pill       & 85    & 91      & 93       & 83.8  & 92.4     & 98.6    & 99.8   & 99.1  & 97.9  & 97.2    &-  & 98.2           \\
Screw      & 96    & 96      & 95       & 84.5  & 86.3     & 97.6    & 97.9   & 99.7  & 97.3  & 98.9     &- & 97.7           \\
Tile       & 51    & 59      & 65       & 98.7  & 93.4     & 100     & 100    & 100   & 99.4  & 98.9     &- & 100            \\
Toothbrush & 93    & 82      & 98       & 100   & 98.3     & 91.9    & 100    & 96.6  & 100   & 100    &-   & 100            \\
Transistor & 86    & 90      & 92       & 90.9  & 95.5     & 99.3    & 92.9   & 99.3  & 100   & 100      &- & 99.65          \\
Wood       & 73    & 73      & 84       & 93    & 98.6     & 100     & 99.5   & 100   & 99.7  & 99.5    &-  & 99.69          \\
Zipper     & 77    & 88      & 87       & 98.1  & 99.4     & 99.7    & 100    & 99.1  & 99.6  & 99.9    &-  & 99.7           \\ \midrule
Avg        & 81.73 & 86.2    & 89.4     & 91.65 & 95.24    & 98.72   & 98.86  & 99.14 & 99.17 & 99.36   & 99.6 & 99.48 \\ \bottomrule
\end{tabular}%
}
\end{table*}
\vspace{-0.2cm}
\subsection{Decoding Module}
\vspace{-0.1cm}
As shown in pink box of \cref{fig:encoder}, decoder network structure consists of $1 \times 1$ convolution, activation and $1 \times 1$ convolution layers. The representation \textbf{R} is divided by spatial resolution, and each spatial feature vector is responsible for the reconstruction of corresponding patch of the original image via $1 \times 1$ convolution. This kind of one-to-one correspondence reconstruction promotes feature vector's ability of capturing fine-grained characteristics of image patch, eliminating the disturbance from other feature vectors. The representation \textbf{R} is input to the decoder to reconstruct the original image \textbf{I}, with the reconstructed result marked as \textbf{\^{I}}.
\vspace{-0.2cm}
\subsection{Loss of Patch AE}
\vspace{-0.2cm}
We train the model with a patch-wise reconstruction loss measured by pair-wise L2 -distance between the reconstructed image \textbf{\^{I}} and its ground truth \textbf{I}. 
\begin{equation}
\setlength{\abovedisplayskip}{2pt} 
\setlength{\belowdisplayskip}{2pt}
\begin{split}
\hat{I}&=\left\{\hat{I}^{1}, \hat{I}^{2}, \cdots, \hat{I}^{p},\cdots,\hat{I}^{P}\right\}_{p=1}^{P} \\ 
&=f^{d}\left[\Gamma_{p=1}^{P}\left(f^{\mathrm{e}}\left(I ; \theta^{\mathrm{e}}\right) \right); \theta^{d}\right]
\end{split}
\label{Eq:reconstructed I}
\end{equation}
In Eq.\ref{Eq:reconstructed I}, $f^{\mathrm{e}}$ denotes encoder, with parameter $\theta^{\mathrm{e}}$. 
$f^{\mathrm{d}}$ denotes decoder, with $\theta^{\mathrm{d}}$ being the parameter of decoder networks. 
$\Gamma_{p=1}^P()$ denotes segmentation function, which divides input into $P$ parts. In Eq.\ref{Eq:reconstructed I}, it divides feature tensor \textbf{R} by spatial resolution. In Eq.\ref{Eq:patched I}, it divides the ground truth image \textbf{I} by patch. 
\begin{equation}
\setlength{\abovedisplayskip}{2pt} 
\setlength{\belowdisplayskip}{2pt}
I=\left\{I^{1}, I^{2}, \cdots,\hat{I}^{p},\cdots, I^{P}\right\}_{p=1}^{P}=\Gamma_{p=1}^{P}(I)
\label{Eq:patched I}
\end{equation}
The whole patch AE model is trained in an end-to-end manner, and the overall loss function is shown as Eq.\ref{Eq:loss all},
\begin{equation}
\setlength{\abovedisplayskip}{2pt} 
\setlength{\belowdisplayskip}{2pt}
\begin{split}
\ell=&\alpha \sum_{p=1}^{P} \|\left(\operatorname{norm}\left(\hat{I}^{p}\right)-\operatorname{norm}\left(I^{p}\right)\right)\|_{2}+\\ 
&(1-\alpha)\sum_{p=1}^{P} \|\hat{I}^{p}-I^{p}\|_{2}
\end{split}
\label{Eq:loss all}
\end{equation}
Pair-wise L2-distance between the reconstructed patch and its ground truth are calculated as reconstruction loss, as Eq.\ref{Eq:loss all}. Inspired by MAE \cite{HeCXLDG22}, In order to obtain a better representation, we do not calculate the loss merely with the pixel value. Furthermore, loss is calculated together with the value normalized within patch, with $\alpha$ being the balanced parameter varying from 0 to 1. Enhancing the contrast locally, hybrid reconstruction loss helps to get high quality representation, which is conducive to anomaly detection.
\vspace{-0.2cm}
\subsection{Calculating Anomaly Score Map}
\vspace{-0.2cm}
After training stage, the representation of normal data from the encoder, that is, normal representation set, is obtained. In the test phase, given a query image, its feature representation is obtained, which is a three dimensional tensor. Each spatially distributed feature vector of the tensor corresponds to a patch of image. For every vector within the feature representation, the distance to the nearest vector in normal representation set is then defined to be its patch-wise anomaly score. The maximum patch-wise anomaly score is its image-wise anomaly score. For fairly comparison with SOTA methods, we adopt the same distance calculation method as Patchcore \cite{RothPZSBG22}.
\vspace{-0.4cm}
\section{Experiments}
\vspace{-0.4cm}
\label{sec:experiments}
Our patch-wise auto-encoder anomaly detection model (Patch AE) is evaluated on MVTec AD dataset on the task of anomaly detection. Patch AE is compared with traditional auto-encoder based model and current state-of-the-art methods in terms of AUROC metric of image level. 

\noindent\textbf{Dataset.} MVTec AD dataset\footnote{http://www.mvtec.com/company/research/datasets} \cite{bergmann2019mvtec} has a total of 15 categories, with 5 of them being distinct types of textures and the remaining 10 being different sorts of objects. In total, 3629 photos are utilized for training, and 1725 images are used for testing in this dataset. The training set contains solely non-defective images, whereas the testing set contains both non-defective and defective images of various types.

\noindent\textbf{Experimental Results.} We compare our approach against traditional auto-encoder based and state-of-the-art unsupervised anomaly detection methods. The results for anomaly detection task on the MVTec AD dataset are listed in Table \ref{table:Experiment result table.}. Patch AE achieves a state-of-the-art overall AUROC score of 99.48\% for anomaly detection, outperforming all of the tested methods. Results demonstrate that model trained with artificial defects \cite{LiSYP21}  generalizes well on unseen and real anomalies. In terms of traditional AE based algorithms, Patch AE exceeds by more than 10\%. In terms of SOTA algorithms, Patch AE outperforms the best recent state-of-the-art method Patchcore 
 on 5 classes and achieves a higher overall AUROC score. Although our model doesn't match the performance of multi-model Patchcore (99.6\%), we are better than its single-model version (99.36\%). In the actual industrial scenario, computing resources are limited and the inference time is required to a certain extent. Our model is more lightweight than multi-model Patchcore. We roughly only use one third computing resources of multi-model Patchcore, which is more promising in the scenario where computation resource is limited but faster inference is required.
\vspace{-0.3cm}
\section{Conclusions}
\vspace{-0.3cm}
\label{sec:conclusions}
In this work, we have primarily presented a general unsupervised anomaly detection approach, i.e. Patch AE. We improve the traditional AE method and propose an idea contrary to typical AE to get a defect sensitive feature representation. This model is realized by patch-wise reconstruction and adding artificial defects for data augmentation. Experiments and analysis on Mvtec AD data have demonstrated that our method is effective and achieves state-of-the-art results. In future work, we plan to further optimize our approach for implementations of multiple model fusion.



\bibliographystyle{IEEEbib}
\bibliography{egbib}
\vspace{-0.4cm}
\end{document}